\documentclass{main}

\usepackage{comment}
\usepackage{subfigure}
\usepackage{amsmath}
\usepackage[normalem]{ulem}
\usepackage{multirow}
\usepackage{times}
\usepackage{epsfig}
\usepackage{graphicx}
\usepackage{amsmath}
\usepackage{amssymb}
\usepackage{booktabs}
\usepackage{comment}
\usepackage{subfigure}
\usepackage{xcolor}
\usepackage{float}
\usepackage{verbatim}

\title{Remote Photoplethysmograph Signal Measurement from Facial Videos Using Spatio-Temporal Networks}

\addauthor{Zitong Yu}{Zitong.Yu@oulu.fi}{1}
\addauthor{Xiaobai Li}{Xiaobai.Li@oulu.fi}{1}
\addauthor{Guoying Zhao\textsuperscript{$^{*}$2,}}{Guoying.Zhao@oulu.fi}{1}

\addinstitution{
 Center for Machine Vision and Signal
Analysis\\
 University of Oulu\\
 FI-90014, Finland
}

\addinstitution{
 School of Information and Technology\\
 Northwest University\\
 710069, China
}

\runninghead{YU, LI, ZHAO}{rPPG MEASUREMENT USING SPATIO-TEMPORAL NETWORKS}

\begin{document}

\maketitle

\vspace{-1em}
\begin{abstract}
Recent studies demonstrated that the average heart rate (HR) can be measured from facial videos based on non-contact remote photoplethysmography (rPPG). However for many medical applications (e.g., atrial fibrillation (AF) detection) knowing only the average HR is not sufficient, and measuring precise rPPG signals from face for heart rate variability (HRV) analysis is needed. Here we propose an rPPG measurement method, which is the first work to use deep spatio-temporal networks for reconstructing precise rPPG signals from raw facial videos. With the constraint of trend-consistency with ground truth pulse curves, our method is able to recover rPPG signals with accurate pulse peaks. Comprehensive experiments are conducted on two benchmark datasets, and results demonstrate that our method can achieve superior performance on both HR and HRV levels comparing to the state-of-the-art methods. We also achieve promising results of using reconstructed rPPG signals for AF detection and emotion recognition.
\end{abstract}


\vspace{-1.5em}
\section{Introduction}
\label{sec:intro}
\vspace{-0.4em}

The heart pulse is an important vital sign that needs to be measured in many circumstances, especially for healthcare or medical purposes. Traditionally, the Electrocardiography (ECG) and Photoplethysmograph (PPG) are the two most common ways for measuring heart activities. From ECG or PPG signals, doctors can get not only the basic average heart rate (HR), but also more detailed information as the inter-beat-interval (IBI) for heart rate variability (HRV) analysis and supporting their diagnosis. However, both ECG and PPG sensors need to be attached to body parts which may cause discomfort and are inconvenient for long-term monitoring. To counter for this issue, remote photoplethysmography (rPPG) is developing fast in recent years, which targets to measure heart activity remotely without any contact.

In earlier studies of rPPG, most methods~\cite{Verkruysse:08,Poh2010,Poh2011,motion2013,Li2014,CHROM,Tulyakov2016,ICME2015} can be seen as a two-stage pipeline, which first detects or tracks the face to extract the rPPG signals, and then estimates the corresponding average HR from frequency analysis. However, there are two disadvantages of these methods that worth concern. 
First, each of them works with self-defined facial regions that are based on pure empirical knowledge but are not necessary the most effective regions, which should vary across data.
Second, the methods involves handcrafted features or filters, which may not generalize well and could lose important information related to heart beat.

\begin{figure}
\centering
\includegraphics[width=12cm,height=2.2cm]{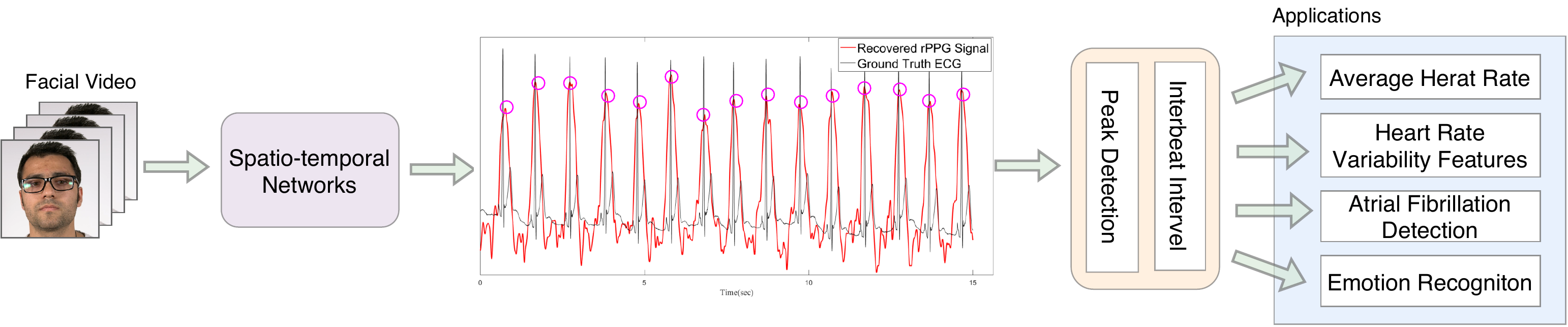}
\vspace{-0.5em}
\caption{\small{Proposed rPPG signal measurement framework using spatio-temporal networks.}}
\label{fig:Figure1}
\vspace{-1em}
\end{figure}


There were also several attempts to estimate HR remotely using deep learning~\cite{IJCB2017,Xuesong2018,HR-CNN,DeepPhys,gang2019Atrial} approaches. But these studies have at least one of the following drawbacks: 
1) The HR estimation task was treated as a one-stage regression problem with one simple output of the average HR, while the individual pulse peak information were lost which limits their usage in demanding medical applications.
2) The approach is not an end-to-end system, which still requires pre-processing or post-processing steps involving handcrafted features. 
3) The approach is based on 2D spatial neural network without considering the temporal context features, which are essential for the rPPG measurement problem.

In this paper, spatio-temporal modeling is conducted on facial videos aiming to locate each individual heartbeat peak accurately. Figure~\ref{fig:Figure1} shows the framework of the proposed rPPG signal measurement method, and steps for related applications. The heartbeat peaks (red circles) of the measured rPPG signal locate precisely at the corresponding R peaks of the ground truth ECG signal, which allows us to achieve not only the average HR, but also detailed IBIs information and HRV analysis for AF detection and emotion recognition.

The main contributions of this work include: 
1) We propose the first end-to-end spatio-temporal network (PhysNet) for rPPG signal measurement from raw facial videos. It takes temporal context into account which was ignored in previous works. 
2) Multiple commonly used spatio-temporal modeling methods are explored and compared, which can serve as a foundation for future network optimization specially for rPPG measurement task. 
3) Compared with state-of-the-art methods, the proposed PhysNet achieves superior performance for measuring not only the average HRs, but also the HRV features, which were further demonstrated to be effective for AF disease detection and emotion recognition.
4) The proposed PhysNet also has good generalized ability on new data as shown in a cross data test. 

\vspace{-1.3em}
\section{Related Work}
\label{sec:related}
\vspace{-0.5em}

Previous methods for remote photoplethysmography measurement, background of HRV measurement, and spatio-temporal networks are briefly reviewed in three subsections.

\vspace{-1.2em}
\subsection{Remote Photoplethysmography (rPPG) Measurement}
\vspace{-0.4em}
In past few years, several traditional methods explored rPPG measurement from videos by analyzing subtle color changes on facial regions of interest (ROI), including blind source separation~\cite{Poh2010, Poh2011}, least mean square~\cite{Li2014}, majority voting~\cite{lam2015robust} and self-adaptive matrix completion~\cite{Tulyakov2016}. There are other traditional methods which utilized all skin pixels for rPPG measurement, e.g., chrominance-based rPPG (CHROM)~\cite{CHROM}, projection plane orthogonal to the skin tone (POS)~\cite{wang2017algorithmic}, and spatial subspace rotation~\cite{2SR}. These methods require complex prior knowledge for ROI selection/skin-pixels detection and handcrafted signal processing steps, which are hard to deploy and do not necessarily generalize well to new data. Besides, the majority of these works only worked on getting the average HR, but did not consider the accuracy of locating each individual pulse peak, which is a more challenging task.

Recently, a few deep learning based methods were proposed for average HR estimation. In~\cite{IJCB2017}, Hsu et al. employed the short-time Fourier transform to build spectrogram and utilized convolutional neural network (CNN) to estimate average HRs. Niu et al.~\cite{Xuesong2018} constructed a spatial-temporal map for CNN to measure average HRs. Radim et al.~\cite{HR-CNN} proposed the HR-CNN, which used the aligned face images to predict HRs. In~\cite{DeepPhys}, Chen and McDuff exploited normalized frame difference for CNN to predict the pulse signal. These methods are not end-to-end frameworks, as they still rely on handcrafted features or aligned face images as inputs. Besides, they were all based on 2D CNN, which lacks the ability to learn the temporal context features of facial sequences which are essential for rPPG signal measurement.

\vspace{-1.2em}
\subsection{Heart Rate Variability Measurement}
\vspace{-0.4em}
Most of the mentioned studies were focusing on average HR measurement. The HR counts the total number of heartbeats in a given time period, which is a very coarse way of describing the cardiac activity. On the other side, HRV features describe heart activity on a much finer scale, which are computed from the IBIs of pulse signals. Most common HRV features include low frequency (LF), high frequency (HF), and their ratio LF/HF, which are widely used in many medical applications. Besides, the respiratory frequency (RF) can also be estimated by analyzing the frequency power of IBI, as in~\cite{Poh2011} and~\cite{OBF}. Apparently, compared with the task of estimating the average HR (only one number), measuring HRV features is more challenging, which requires accurate measure of the time location of each individual pulse peak. For the needs of most healthcare applications, average HR is far from enough. We need to step forward to develop methods that can measure heart activity on HRV level.

\vspace{-1.2em}
\subsection{Spatio-temporal Networks}
\vspace{-0.4em}

Spatio-temporal network plays a crucial role in many video-based tasks (e.g., action detection and recognition~\cite{yang2017spatio}) because of the excellent performance. There are two mainstreams of spatio-temporal frameworks. The first category includes 3D convolutional neural networks (3DCNN) such as Convolutional 3D~\cite{C3D}, Pseudo 3D~\cite{P3D}, Inflated 3D~\cite{carreira2017quo} and Separable 3D~\cite{xie2018rethinking}, which are widely used for video understanding as they can capture spatial and temporal context simultaneously. The second category includes recurrent neural network (RNN) based frameworks, such as long short term memory (LSTM)~\cite{hochreiter1997long} and Convolutioal LSTM~\cite{xingjian2015convolutional}, which can also capture the temporal context among the CNN spatial features.

Existing spatio-temporal networks are mostly designed for analyzing large scale motions, and it is still unknown whether they are suitable for the rPPG signal measurement task as the temporal skin color variation is extremely subtle. In this paper various spatio-temporal modeling methods and loss functions are evaluated, which will serve as a foundation for future works about network optimization specialized for the rPPG measurement task.


\vspace{-1.2em}
\section{Methodology}
\label{sec:approach}
\vspace{-0.4em}

In this section, the proposed method with several alternative spatio-temporal models is demonstrated. We also describe our customized loss function and rationalize the design. 

\begin{figure}
\centering
\includegraphics[width=12cm,height=5.1cm]{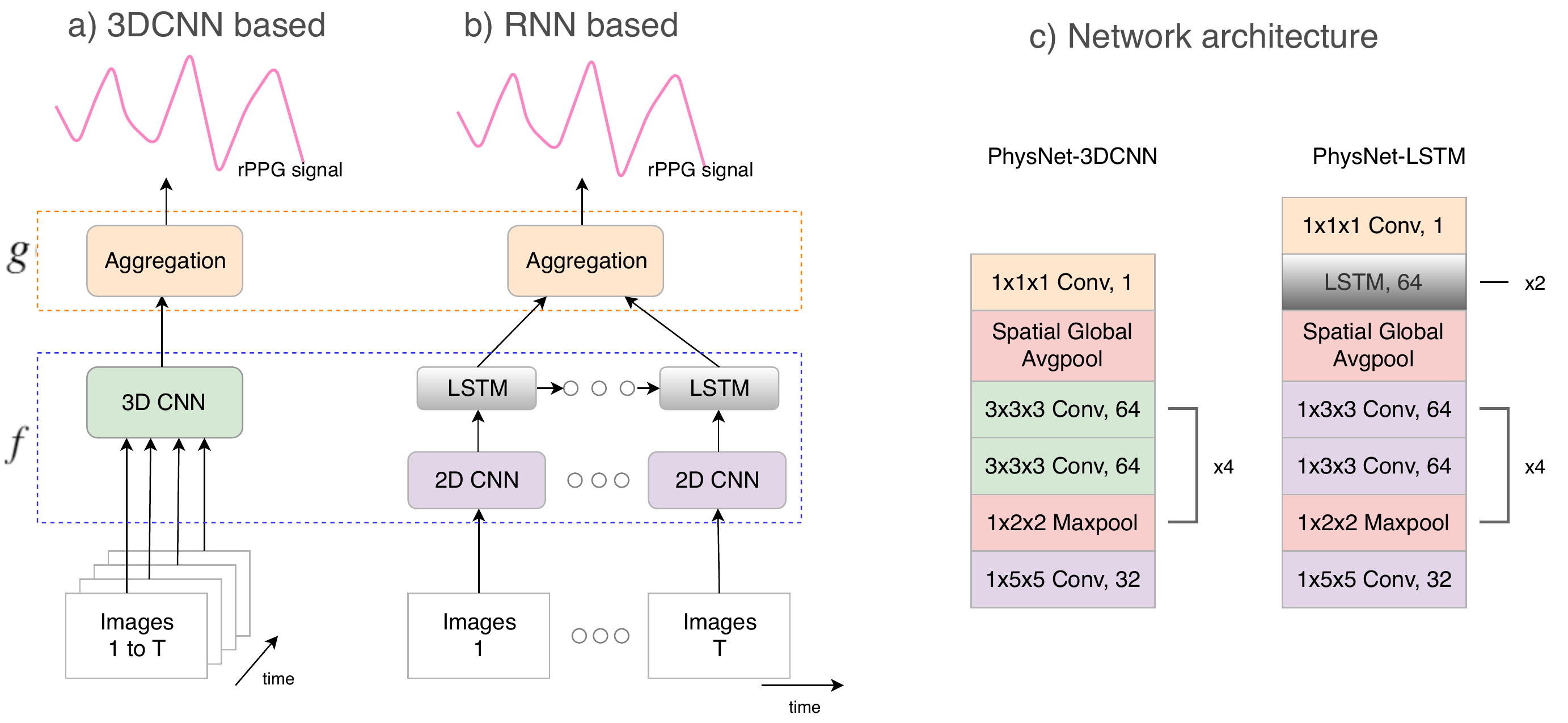}
\vspace{-0.8em}
\caption{\small{The framework of saptio-temporal networks for rPPG signal recovery. a) 3DCNN based PhysNet; b) RNN based PhysNet; c) Their corresponding network architectures. "3x3x3 Conv, 64" donotes using convolution filter with kernel $3\times 3\times 3$ and output channel number is 64 while other operations are analogous.}}
\label{fig:framework}
\vspace{-1.2em}
\end{figure}

\vspace{-1.2em}
\subsection{Network Architecture}
\vspace{-0.4em}
\label{sec:Network}

According to~\cite{CHROM, wang2017algorithmic}, there are two important procedures in order to achieve pulse information from facial videos. First is to project RGB into color subspace with stronger representation capacity. After that, the color subspace needs to be reprojected in order to get rid of irrelevant info (e.g., noise caused by illumination or motion) and achieve the target signal space. Here, we propose an end-to-end spatio-temporal network (denoted as PhysNet), which is able to merge these two steps and achieve the final rPPG signals in one step.


The overall architecture of PhysNet is shown in Figure~\ref{fig:framework}. The input of the network is $T$-frame face images with RGB channels. After forwarding several convolution and pooling operations, multi-channel manifolds are formed to represent the spatio-temporal features. Finally, the latent manifolds are projected into signal space using channel-wise convolution operation with \(1\times1\times1\) kernel to generate the predicted rPPG signal with length $T$. The whole procedure can be formulated as
\vspace{-0.6em}
\begin{eqnarray}
[y_{1},y_{2},...,y_{T}]=g(f([x_{1},x_{2},...,x_{T}];\theta);w),
\end{eqnarray}
\vspace{-1.5em}

\noindent where \([x_{1},x_{2},...,x_{T}]\) are the input frames, and \([y_{1},y_{2},...,y_{T}]\) is the output signal of the network. As shown in Figure~\ref{fig:framework}(a)(b), \(f\) is the spatio-temporal model for subspace projection, \(\theta\) is a concatenation of all convolutional filter parameters of this model, \(g\) is the channel aggregation for final signal projection, and \(w\) is a set of its parameters. 
For the spatio-temporal model \(f\) there are two mainstream models, and here we explore and compare both, as 3DCNN based and RNN based PhysNet.

\vspace{0.4em}

\noindent\textbf{3DCNN based PhysNet}  \quad    Denoted as "PhysNet-3DCNN" and shown in Figure~\ref{fig:framework}(a)(c), 3DCNN is imposed as the spatio-temporal model \(f\), which adopts \(3\times3\times3\) convolutions to extract the semantic rPPG features in both spatial and temporal domain simultaneously. It helps to learn more robust context features and recover the rPPG signals with less temporal fluctuation. 
Inspired by successful leveraging encoder-decoder in action segmentation task~\cite{lea2017temporal}, we also attempt a temporal encoder-decoder (ED) structure for rPPG task, denoted as "PhysNet-3DCNN-ED", which intends to exploit more effective temporal context and reduce temporal redundancy and noise. 


\vspace{0.4em}

\noindent\textbf{RNN based PhysNet} \quad      In Figure~\ref{fig:framework}(b)(c), 2DCNN is deployed to extract the spatial features firstly and then RNN based module is exploited for propagating the spatial features in temporal domain, which may improve the temporal context features via forward/backward information flows. LSTM and ConvLSTM can be formulated as
\vspace{-0.1em}
\begin{equation} 
\begin{split}
&i_{t}=\delta (W_{i}^{X}\ast X_{t}+W_{i}^{H}\ast H_{t-1}), \\
&f_{t}=\delta (W_{f}^{X}\ast X_{t}+W_{f}^{H}\ast H_{t-1}), \\
&o_{t}=\delta (W_{o}^{X}\ast X_{t}+W_{o}^{H}\ast H_{t-1}), \\
&c_{t}=f_{t}\circ c_{t-1}+i_{t}\circ tanh(W_{c}^{X}\ast X_{t}+W_{c}^{H}\ast H_{t-1}), \\
&H_{t}=o_{t}\circ tanh(c_{t}), \\
\end{split}
\vspace{-0.2em}
\end{equation}
where $\ast $ denotes the multiplication and convolution operator for LSTM and ConvLSTM respectively, and $\circ $ denotes the Hadamard product. Here, bias terms are omitted. All the gates $i$, $f$, $o$, memory cell $c$, hidden state $H$ and the learnable weights $W$ are 3D tensors. In later sections, "PhysNet-LSTM", "PhysNet-BiLSTM" and "PhysNet-ConvLSTM" represent the networks with LSTM, bi-directional LSTM and Convolutional LSTM respectively. For PhysNet-ConvLSTM, global average pooling is deployed after temporal propagation.

About implementation details, all the convolution operations use \(1\times1\times1\) stride, and are cascaded with batch normalization and nonlinear activation function ReLU except the last convolution with a \(1\times1\times1\) kernel. The strides of all the "Maxpool" operations are set as $1\times2\times2$ except "PhysNet-3DCNN-ED". For "PhysNet-3DCNN-ED", both strides and kernel sizes of the second and third "Maxpool" are $2\times2\times2$ in encoder while there are two deconvolution layers before "Spatial Global Avgpool" in decoder to reach back to the original temporal length. In addition, spatial and temporal padding for all convolutions are needed to keep a consistent size. The number of the stacked LSTM layers are set as 2. As our method is designed as a fully convolutional framework so theoretically facial sequences with arbitrary spatial and temporal size are feasible as inputs.

\vspace{-1.2em}
\subsection{Loss Function}
\vspace{-0.4em}
\label{sec:Loss}

Besides designing the network architecture, we also need an appropriate loss function to guide the networks. One major step-forward of the current study is that we aim to recover rPPG signals which have matching trend and accurately estimated pulse peak time locations that match with ground truth signals, which are essential for detailed HRV analysis. In order to maximize the trend similarity and minimize peak location errors, negative Pearson correlation is utilized as the loss function

\vspace{-0.6em}
\begin{equation} \small
Loss=1-\frac{T\sum_{1}^{T}xy-\sum_{1}^{T}x\sum_{1}^{T}y}{\sqrt{(T\sum_{1}^{T}x^2-(\sum_{1}^{T}x)^2)(T\sum_{1}^{T}y^2-(\sum_{1}^{T}y)^2)}},
\vspace{-0.3em}
\end{equation}
where \(T\) is the length of the signals, \(x\) is the predicted rPPG signals, and \(y\) indicates the ground truth PPG signals.

PPG signals are used as the ground truth for training our network instead of ECG because PPG measured from fingers resembles more to the rPPG measured from faces, as they both measure the peripheral blood volume changes, while ECG measures electrical activities thus contains extra components that do not present in rPPG. In the testing stage, ECG is adopted as the ground truth following previous works like~\cite{Xuesong2018,HR-CNN,DeepPhys} for fair comparison.

\vspace{-1.5em}
\section{Experiments}
\label{sec:Experiments}
\vspace{-0.8em}

Two datasets are employed in our experiments. First, we train the proposed PhysNet on the OBF dataset~\cite{OBF}. OBF has large number of facial videos and with corresponding PPG signals, which are suited for our training need. The trained PhysNet is first tested on OBF for evaluation of HR and HRV measurement accuracy, and then demonstrated for an extended application of AF detection. At last the trained PhysNet is also crossly tested on the MAHNOB-HCI~\cite{MAHNOB} dataset and another application of emotion recognition is explored.

\vspace{-1.3em}
\subsection{Data and Experimental Settings}
\vspace{-0.4em}

\textbf{OBF dataset}~\cite{OBF} is used for both training and testing. OBF contains totally 212 videos recorded from 100 healthy adults and six patients with atrial fibrillation (AF). Each subject were recorded for two five-minute sessions, in which facial videos and the corresponding  physiological signals (two video cameras, ECG and breathing signal measured from chest, and PPG from finger) were recorded simultaneously. In the current study we use the RGB videos, which were recorded at 60 fps with resolution of 1920x2080.

\noindent\textbf{MAHNOB-HCI dataset}~\cite{MAHNOB} is used for cross testing the generalization of our model. It includes 527 videos from 27 subjects, with recording speed of 61 fps and resolution of 780x580. We use the EXG2 signals as the ground truth ECG signal in evaluation. In order to make fair comparison with previous works~\cite{Xuesong2018,HR-CNN,DeepPhys}, we follow the same routine as their works and use 30 seconds clip (frames 306 to 2135) of each video.

\vspace{0.5em}

\noindent\textbf{Training Settings.}\quad Facial videos and corresponding PPG signals are synchronized before training. For each video clip, we use the Viola-Jones face detector ~\cite{VJface} to crop the face area at the first frame and fix the region through the following frames. Then the face images are normalized to 128x128. We set the length of training clips as $T=\left \{ 32,64,128,256 \right \}$ and both videos and ground truth signals are downsampled to 30 fps and 30 Hz respectively. The proposed method is trained on Nvidia P100 with PyTorch. The Adam optimizer is used and the learning rate is set as 1e-4. We train all the models for 15 epochs.

\vspace{0.5em}






\noindent\textbf{Testing Settings and Performance Metrics.}\quad  
As PhysNet is designed as fully-convolutional structure, it is easily to do inference in arbitrary long video clips. In the testing stage, both the recovered rPPGs and their corresponding ground truth ECG signals go through the same process of filtering, normalization, and peak detection to obtain the inter-beat-intervals, from which the average HR and HRV features are calculated. For HRV level evaluation, we followe paper~\cite{Poh2011} to calculate three commonly used HRV features (in normalized units, n.u.) and the RF (in Hz). Details about the features are referred to~\cite{Poh2011,OBF}. Performance metrics for evaluating both the average HR and HRV features include: the standard deviation (SD), the root mean square error (RMSE), the Pearson's correlation coefficient (R), and the mean absolute error (MAE). For atrial fibrillation detection and emotion recognition evaluation, the accuary (ACC) and specificity (SP) are used as the validation metrics.

\vspace{-1em}
\subsection{Experiments on OBF}
\vspace{-0.3em}
We evaluate several aspects of the proposed method separately, i.e., the loss function, spatio-temporal networks and clip length, and report performance on both HR and HRV levels. Subject-independent 10-fold cross validation is adopted here. We also report the AF detection accuracy of using the measured HRV features as an application scenario.

\vspace{0.5em}
\noindent\textbf{Loss Function.} \quad   
In order to demonstrate the advantages of our proposed negative Pearson (NegPea) loss function, we compare it with the mean square error (MSE), which is used in~\cite{DeepPhys}. Both experiments employed 3DCNN based PhysNet with the training clip length fixed to 64 (as PhysNet64-3DCNN), and the results are listed in Table~\ref{tab:lossfunction}. Results show that NegPea performs better than MSE on both HV and HRV levels, which support the efficacy of the proposed loss function as we explained in Sec~\ref{sec:Loss}. Note that as the amplitude of peaks are irrelevant with our task (i.e., to measure accurate time location of heartbeats), the MSE loss may direct wrongly and introduces extra noises.


\begin{figure}[H]
\vspace{-1em}
\centering
\includegraphics[width=13.2cm,height=2.8cm]{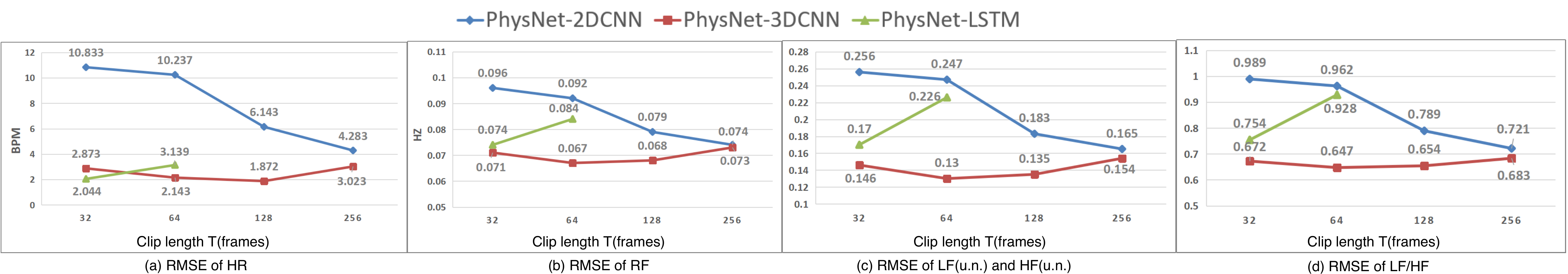}
\vspace{-1.8em}
\caption{\small{Performance comparison of various training clip lengths $T=32,64,128,256$. RMSE is used as the evaluation metric, lower RMSE value indicates better performance.}}
\label{fig:cliplength}
\vspace{-1.4em}
\end{figure}

\vspace{0.5em}
\noindent\textbf{Spatio-temporal Networks.} \quad   In these experiments we evaluate the effectiveness of spatio-temporal networks, and we fixed the training clip length as $T=64$ and NegPea as the loss function. 
First, we evaluated a 2DCNN based PhysNet (PhysNet-2DCNN) and reported its result as the baseline (Table~\ref{tab:STNetworks} top).
Second, we evaluate 3DCNN based models either with (PhysNet64-3DCNN-ED) or without (PhysNet64-3DCNN) the ED component(Table~\ref{tab:STNetworks} middle).
It is clear that 1) compared with 2DCNN, temporal convolutions boost performance on both HR and HRV levels, and 2) benefited by the temporal encoder-decoder structure, "PhysNet64-3DCNN-ED" achieved better performance than "PhysNet64-3DCNN". It can be explained that semantic features with less temporal redundancy can be extracted in such




\noindent squeeze-stretch-like encoding and decoding process. Third, we also evaluate how RNN based models perform, i.e., using LSTM (PhysNet64-LSTM), Bidirectional LSTM (PhysNet64-BiLSTM) and Convolutional LSTM (PhysNet64-ConvLSTM). Results (Table~\ref{tab:STNetworks} bottom) show that 1) "PhysNet64-LSTM" achieved better performance than the baseline "PhysNet64-2DCNN" but not as well as 3DCNN, which implies that LSTM is able to improve the performance but not so effective as 3DCNN for long-term temporal context aggregation; and 
2) LSTM and ConvLSTM are about the same level while BiLSTM is the worst, which indicates the backward information of the highest-level features seems to be not necessary.

\begin{table*}[t]\footnotesize
\begin{center}
\caption{\small{Performance comparison of two loss functions with negative Pearson and MSE. Smaller RMSE and bigger R values indicate better performance.}}
\vspace{-0.6em}
\label{tab:lossfunction}
\begin{tabular}{p{3.4cm}  p{0.5cm} p{0.5cm}  p{0.5cm} p{0.5cm}   p{0.5cm} p{0.5cm}  p{0.5cm} p{0.5cm}  p{0.5cm} p{0.5cm}}

\toprule
& \multicolumn{2}{c}{HR(bpm)} &  \multicolumn{2}{c}{RF(Hz)} &  \multicolumn{2}{c}{LF(u.n)} &  \multicolumn{2}{c}{HF(u.n)} &  \multicolumn{2}{c}{LF/HF}\\
  
\cmidrule(lr){2-3} \cmidrule(lr){4-5}
\cmidrule(lr){6-7} \cmidrule(lr){8-9}
\cmidrule(lr){10-11}

  \multicolumn{1}{c}{Method} & RMSE & \multicolumn{1}{c}{R}  &  RMSE & \multicolumn{1}{c}{R}  & RMSE & \multicolumn{1}{c}{R}  & RMSE & \multicolumn{1}{c}{R}  & RMSE & \multicolumn{1}{c}{R}\\

 \midrule
 PhysNet64-3DCNN-MSE   & 4.012 & 0.955    & 0.069  & 0.435   & 0.169 & 0.689  & 0.169 & 0.689  & 0.721 & 0.659\\
 
  PhysNet64-3DCNN-NegPea  & \underline{2.143} & \underline{0.985}   & \underline{0.067}  & \underline{0.494}  & \underline{0.15} & \underline{0.749}  & \underline{0.15} & \underline{0.749} & \underline{0.647} & \underline{0.72}\\

\bottomrule
\end{tabular}
\end{center}
\vspace{-2em}

\end{table*}

\begin{table*}[t]\footnotesize
\begin{center}
\caption{\small{Performance comparison of spatio-temporal networks.}}
\vspace{-0.6em}
\label{tab:STNetworks}
\begin{tabular}{p{2.9cm}  p{0.55cm} p{0.55cm}  p{0.55cm} p{0.55cm}   p{0.55cm} p{0.55cm}  p{0.55cm} p{0.55cm}  p{0.55cm} p{0.55cm}}

\toprule
& \multicolumn{2}{c}{HR(bpm)} &  \multicolumn{2}{c}{RF(Hz)} &  \multicolumn{2}{c}{LF(u.n)} &  \multicolumn{2}{c}{HF(u.n)} &  \multicolumn{2}{c}{LF/HF}\\
  
\cmidrule(lr){2-3} \cmidrule(lr){4-5}
\cmidrule(lr){6-7} \cmidrule(lr){8-9}
\cmidrule(lr){10-11}

   \multicolumn{1}{c}{Method} & RMSE & \multicolumn{1}{c}{R}  &  RMSE & \multicolumn{1}{c}{R}  & RMSE & \multicolumn{1}{c}{R}  & RMSE & \multicolumn{1}{c}{R}  & RMSE & \multicolumn{1}{c}{R}\\

 \midrule
 PhysNet64-2DCNN   & 10.237 & 0.928    & 0.092  & 0.104   & 0.247 & 0.321  & 0.247 & 0.321  & 0.962 & 0.318\\
 
 \midrule
 
 

 PhysNet64-3DCNN & 2.143 & 0.985   & 0.067  & 0.494  & 0.15 & 0.749  & 0.15 & 0.749 & 0.647 & 0.72\\
 
  PhysNet64-3DCNN-ED & \underline{2.048} & \underline{0.989}   & \underline{0.066}  & \underline{0.501}  & \underline{0.146} & \underline{0.772}  & \underline{0.146} & \underline{0.772} & \underline{0.624} & \underline{0.748}\\
 
 \midrule
 
 PhysNet64-LSTM   & 3.139 & 0.975   & 0.084  & 0.189 & 0.226 & 0.478 & 0.226 & 0.478  & 0.928 & 0.404\\
 
   PhysNet64-BiLSTM   & 4.595 & 0.945  & 0.085  & 0.183 & 0.231 & 0.421  & 0.231 & 0.421 & 0.956 & 0.396\\

 PhysNet64-ConvLSTM   & 2.937 & 0.977   & 0.083  & 0.191 & 0.22 & 0.485  & 0.22 & 0.485 & 0.896 & 0.44\\

\bottomrule
\end{tabular}
\end{center}
\vspace{-2.3em}

\end{table*}

\begin{table*}[t]\footnotesize
\begin{center}

\caption{\small{Performance comparison between previous methods and our proposed method.}}
\vspace{-0.8em}
\label{tab:Previous}
\begin{tabular}{p{2.9cm}  p{0.55cm} p{0.55cm}  p{0.55cm} p{0.55cm}   p{0.55cm} p{0.55cm}  p{0.55cm} p{0.55cm}  p{0.55cm} p{0.55cm}}

\toprule
& \multicolumn{2}{c}{HR(bpm)} &  \multicolumn{2}{c}{RF(Hz)} &  \multicolumn{2}{c}{LF(u.n)} &  \multicolumn{2}{c}{HF(u.n)} &  \multicolumn{2}{c}{LF/HF}\\
  
\cmidrule(lr){2-3} \cmidrule(lr){4-5}
\cmidrule(lr){6-7} \cmidrule(lr){8-9}
\cmidrule(lr){10-11}

   \multicolumn{1}{c}{Method} & RMSE & \multicolumn{1}{c}{R}  &  RMSE & \multicolumn{1}{c}{R}  & RMSE & \multicolumn{1}{c}{R}  & RMSE & \multicolumn{1}{c}{R}  & RMSE & \multicolumn{1}{c}{R}\\

 \midrule
 ROI\_green~\cite{OBF}   & 2.162 & 0.99    & 0.084  & 0.321   & 0.24 & 0.573  & 0.24 & 0.573  & 0.832 & 0.571\\
 
 CHROM~\cite{CHROM}    & 2.733 & 0.98   & 0.081  & 0.224  & 0.206 & 0.524  & 0.206 & 0.524  & 0.863 & 0.459\\
 
 POS~\cite{wang2017algorithmic}    & 1.906 & 0.991   & 0.07  & 0.44   & 0.158 & 0.727  & 0.158 & 0.727 & 0.679 & 0.687\\
 
 \midrule

 PhysNet128-3DCNN-ED  & \textbf{1.812} & \textbf{0.992}   & \textbf{0.066}  & \textbf{0.507}  & \textbf{0.148} & \textbf{0.766}  & \textbf{0.148} & \textbf{0.766} & \textbf{0.631} & \textbf{0.739}\\
\bottomrule
\end{tabular}
\end{center}
\vspace{-2em}

\end{table*}

\vspace{0.5em}
\noindent\textbf{Clip Length $T$.} \quad   In training stage, the video length may impact each network differently, and we evaluate $T=\left \{ 32,64,128,256 \right \}$ here. Results are shown in Figure~\ref{fig:cliplength}. For "PhysNet-2DCNN" it is clear that longer inputs lead to better performance. 
"PhysNet-3DCNN" has more stable performance over clip length. Note that with temporal convolution layers, "PhysNet32-3DCNN" outperformed "PhysNet-2DCNN" with much shorter inputs, as the temporal convolution filters can provide extra help in learning temporal representations. 
For "PhysNet-LSTM" we only compared $T=\left \{ 32,64 \right \}$ because of its limited long-term temporal propagation ability, and $T=32$ achieved better performance. For "PhysNet-3DCNN", the best HR and HRV performance were achieved at $T=128$ and $T=64$ respectively.

\vspace{0.5em}
\noindent\textbf{Comparison with Previous Methods.}\quad    We replicate three previous methods, i.e.,~\cite{OBF} as "ROI\_green", ~\cite{CHROM} as "CHROM", and~\cite{wang2017algorithmic} as "POS", and compare with our method in Table~\ref{tab:Previous}. Our best performance was achieved with "PhysNet128-3DCNN-ED" (marked in bold), which outperforms all compared methods on both HR and HRV levels indicating efficacy and robustness of the proposed method.


\vspace{0.5em}
\noindent\textbf{Atrial Fibrillation Detection.}\quad    Followed the protocol in~\cite{OBF}, we extracted ten dimensional HRV features from the recovered rPPG signals for detecting AF cases against healthy ones. As seen in Table~\ref{tab:ResultsAF}, results show that PhysNet achieves better performance than previous methods. 
Note that the pre-trained parameters of classifers were fixed and kept same, so the improvement was purely based on the more accurately measured HRV features by PhysNet.

\begin{table}\footnotesize
\vspace{-0.1em}
\centering
\caption{\small{Results of Atrial Fibrillation Detection on OBF.}}
\label{tab:ResultsAF}
 \begin{tabular}{l c c c c c} 
 \toprule
 Metric & ROI\_green~\cite{OBF} & CHROM~\cite{CHROM}& POS~\cite{wang2017algorithmic} & PhysNet128-3DCNN-ED (ours)  \\
 \midrule
 ACC & 77.23\% & 70.61\% & 76.5\% & \textbf{80.22\%} \\ 
 SP & 75.61\% & 74.18\% & 79.37\% & \textbf{81.71\%} \\
 \bottomrule
 \end{tabular}
\vspace{-1em}
\end{table}

\begin{table}\small
\vspace{-1.2em}
\centering
\caption{\small{Results of average HR measurement on MAHNOB-HCI.}}
\label{tab:ResultsMAHNOB}
 \begin{tabular}{l c c c c} 
 \toprule
 Method & HR$_{SD}$ & HR$_{MAE}$ & HR$_{RMSE}$ & HR$_{R}$\\
 & (bpm) & (bpm) & (bpm) & \\
 \midrule
 Poh2011~\cite{Poh2011} & 13.5 & - & 13.6 & 0.36 \\ 
 CHROM~\cite{CHROM} & - & 13.49 & 22.36 & 0.21 \\
 Li2014~\cite{Li2014} & 6.88 & - & 7.62 & 0.81\\
 SAMC~\cite{Tulyakov2016} & 5.81 & - & 6.23 & 0.83\\
 SynRhythm~\cite{Xuesong2018} & 10.88 & - & 11.08 & - \\ 
 HR-CNN~\cite{HR-CNN} & - & 7.25 & 9.24 & 0.51 \\
 DeepPhys~\cite{DeepPhys} & - & 4.57 & - & -\\
 
  \midrule
 PhysNet128-3DCNN-ED (ours) & 7.84 & 5.96 & 7.88 & 0.76\\
 \bottomrule
 \end{tabular}
 \vspace{-1em}
\end{table}

\vspace{-1em}
\subsection{Evaluation on MAHNOB}
\vspace{-0.3em}

As "PhysNet128-3DCNN-ED" achieved the best performance on OBF, we use the model trained on OBF to cross test on MAHNOB-HCI to validate its generalization ability. Average HR results of our method are compared with previous methods in Table~\ref{tab:ResultsMAHNOB} (as previous works only reported performance on average HR level but not on HRV level). 
The first four~\cite{Poh2011,CHROM,Li2014,Tulyakov2016} are earlier methods not involving neural network. Although performance of~\cite{Li2014} and~\cite{Tulyakov2016} are good, the approaches are not trained-ready but require computational costly processing steps for each input, which are limited for real-time usage. On the other side the proposed method is a pre-trained end-to-end system which runs very fast on new test samples (discussed later in Sec~\ref{sec:Speed}).
The later three~\cite{Xuesong2018,HR-CNN,DeepPhys} are all neural network based methods, but note that their test protocols vary (e.g.,~\cite{Xuesong2018}(Table 4) trained on MAHNOB-HCI,~\cite{HR-CNN} trained on a lot more (2302) videos from both the whole MAHNOB and another dataset PURE~\cite{stricker2014non}, and~\cite{DeepPhys} trained on other self-collected data), so the results are compared on a general level. Another fact is that all three approaches need pre-processing steps but ours does not thus easy and efficient to deploy. As being cross tested, our method achieved top level performance which indicates its generalization ability for HR measurement.



\vspace{0.4em}
\noindent\textbf{Emotion Recognition.}\quad    Another advantage of the proposed method is that the recovered rPPG signals allow HRV feature analysis for more sophisticated applications, i.e., emotion recognition, while previous methods that estimate only HR values are not feasible.
We extracted ten dimensional HRV features (the same as in AF detection task, see~\cite{OBF}) from rPPG signals measured with "PhysNet128-3DCNN-ED" on MAHNOB-HCI clips, and feed them to a support vector machine (with a polynomial kernel) as the classifier for estimating the person's emotion status of each video. 
Several emotion labels are provided by MAHNOB-HCI, and we follow ~\cite{huang2016multi} to estimate "Arousal" and "Valence" on three levels, and "Emotions" in nine categories.
As listed in Table~\ref{tab:ResultsEmotion}, the results are very promising especially for the valence recognition.
To the best of our knowledge this is the first exploration of using face video measured physiological features for emotion analysis. In next work we will try fusing it with facial expression analysis for multimodal emotion recognition.




\begin{table}\footnotesize
\vspace{-0.1em}
\centering
\caption{\small{Accuracy results of Emotion Recognition on MAHNOB-HCI.}}
\label{tab:ResultsEmotion}
 \begin{tabular}{l c c c c c} 
 \toprule
 Method & Valence (3 classes) & Arousal (3 classes) & Emotion (9 classes)  \\
 \midrule
 PhysNet128-3DCNN-ED & 46.86\% & 44.02\% & 29.79\%\\ 
 \bottomrule
 \end{tabular}
 \vspace{-1em}
\end{table}

\begin{figure}
\centering
\includegraphics[width=9cm,height=3.3cm]{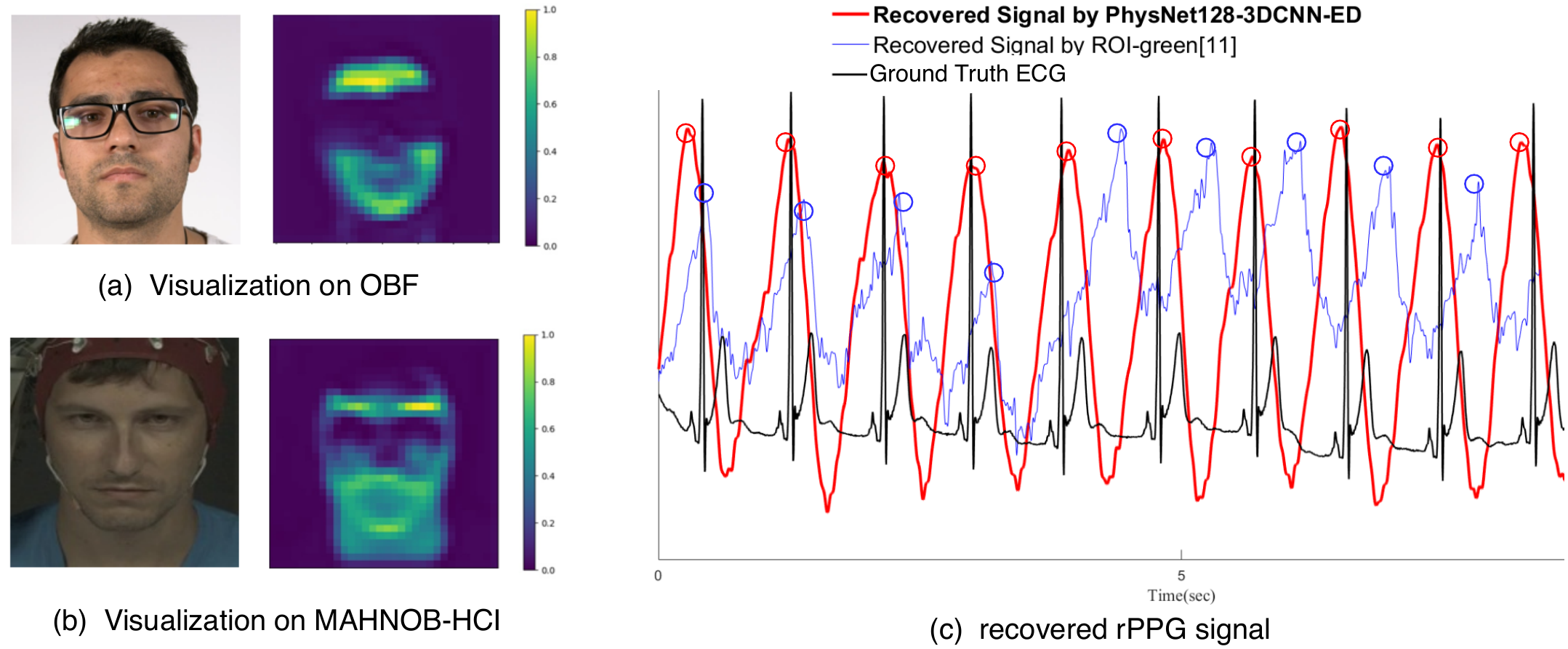}
\vspace{-1em}
\caption{\small{Visualization of original faces, learned neural features and recovered rPPG signals.}}
\label{fig:visualization}
\vspace{-1.2em}
\end{figure}

\vspace{-1.4em}
\subsection{Visualization and Inference Speed}
\label{sec:Speed}
\vspace{-0.4em}
\noindent\textbf{Visualization.}\quad    
Mid-level neural features extracted from both dataset samples by "PhysNet\\128-3DCNN-ED" are shown in Fig.~\ref{fig:visualization} (a) and (b). The high light areas show that the network is able to learn and select skin regions with the strongest rPPG infomation (e.g., forehead, cheeks and lower jaw). Besides, in Fig.~\ref{fig:visualization} (c) we also show a sample rPPG signal recovered with the proposed PhysNet (red) comparing with that from a baseline method "ROI\_green"~\cite{OBF} (blue) and the ground truth ECG (black). The red curve matches much better to the ground truth than the blue one in terms of peak time locations, which demonstrates the effectiveness of the proposed method.



\vspace{0.4em}
\noindent\textbf{Inference Speed.}\quad   As our method does not require any pre-processing step as previous networks~\cite{Xuesong2018} did, it works faster and allow real-time rPPG signal recovery. For a test video of 30s, the "PhysNet64-3DCNN-ED" takes only 0.235s (3830 fps) on a Tesla P100 GPU, which suits most real-time applications.

\vspace{-1em}
\section{Conclusion}
\vspace{-0.7em}
\label{sec:Conclusion and Future Work}
In this paper, we proposed an end-to-end framework with spatio-temporal networks which is able to recover rPPG signals from raw facial videos fast and efficiently. We tested on OBF and MAHNOB-HCI datasets, and results showed that the proposed PhysNet can recover rPPG signals with accurate time location of each individual pulse peak, which allows measuring not only the average HRs, but also IBIs information and HRV level features that enable potential applications in e.g., remote AF detection and emotion recognition.

\vspace{-1.5em}
\section{Aknowledgement}
\vspace{-0.7em}
This work was supported by the National Natural Science Foundation of China (No. 61772419), Tekes Fidipro Program
(No. 1849/31/2015), Business Finland Project (No. 3116/31/2017), Academy of Finland, and Infotech Oulu. As well, the authors wish to acknowledge CSC-IT Center for Science, Finland, for computational resources.

\bibliography{egbib}
\end{document}